\documentclass[a4paper,fleqn]{cas-sc}

\usepackage[authoryear]{natbib}
\usepackage{utfsym}
\usepackage{tabularx}
\usepackage{upgreek}
\usepackage{algorithm}
\usepackage{algpseudocode}
\usepackage{lineno}

\def\tsc#1{\csdef{#1}{\textsc{\lowercase{#1}}\xspace}}
\tsc{WGM}
\tsc{QE}

\begin{document}
\let\WriteBookmarks\relax
\def\floatpagepagefraction{1}
\def\textpagefraction{.001}

\shorttitle{}    

\shortauthors{W.Zhang et al.}  

\title [mode = title]{A Resource-Efficient Training Framework for Remote Sensing Text--Image Retrieval}  



%

\author[1, 3]{Weihang Zhang}
\ead{zhangweihang21@mails.ucas.ac.cn}
\credit{Conceptualization of this study, Methodology, Software}
\author[1,2]{Jihao Li}
\ead{lijihao17@mails.ucas.edu.cn}
\credit{supervision, writing}
\author[1,2]{Shuoke Li}
\ead{lisk@aircas.ac.cn}
\author[3,4]{Ziqing Niu}
\ead{niuziqing21@mails.ucas.ac.cn}
\author[1,2]{Jialiang Chen}
\ead{chenjl@aircas.ac.cn}
\author[1,3]{Wenkai Zhang}
\ead{zhangwk@aircas.ac.cn}
\cormark[1]
\cortext[1]{Corresponding author. This work was supported by the National Key R\&D Program of China under grant 2022ZD0118402.}
\author[1,2]{Xin Gao}
\ead{gaoxin@aircas.ac.cn}
\author[1,2]{Xian Sun}
\ead{sunxian@aircas.ac.cn}



\affiliation[1]{organization={Aerospace Information Research Institute, Chinese Academy of Sciences},
            city={Beijing},
            postcode={100190},
            country={China}}
            
\affiliation[2]{organization={Key Laboratory of Network Information System Technology (NIST), Aerospace Information Research Institute, Chinese Academy of Sciences},
            city={Beijing},
            postcode={100190},
            country={China}}
\affiliation[3]{organization={University of Chinese Academy of Sciences},
            city={Beijing},
            postcode={100190},
            country={China}}
\affiliation[4]{organization={School of Electronic, Electrical and Communication Engineering, University of Chinese Academy of Sciences},
            city={Beijing},
            postcode={100190},
            country={China}}


%


\begin{abstract}
Remote sensing text--image retrieval (RSTIR) aims to retrieve the matched remote sensing (RS) images from the database according to the descriptive text. Recently, the rapid development of large visual-language pre-training models provides new insights for RSTIR. Nevertheless, as the complexity of models grows in RSTIR, the previous studies suffer from suboptimal resource efficiency during transfer learning. To address this issue, we propose a computation and memory-efficient retrieval (CMER) framework for RSTIR. 
To reduce the training memory consumption, we propose the Focus-Adapter module, which adopts a side branch structure. Its focus layer suppresses the interference of background pixels for small targets. 
Simultaneously, to enhance data efficacy, we regard the RS scene category as the metadata and design a concise augmentation technique. The scene label augmentation leverages the prior knowledge from land cover categories and shrinks the search space. We propose the negative sample recycling strategy to make the negative sample pool decoupled from the mini-batch size. It improves the generalization performance without introducing additional encoders. 
We have conducted quantitative and qualitative experiments on public datasets and expanded the benchmark with some advanced approaches, which demonstrates the competitiveness of the proposed CMER. Compared with the recent advanced methods, the overall retrieval performance of CMER is 2\%--5\% higher on RSITMD. Moreover, our proposed method reduces memory consumption by 49\% and has a 1.4x data throughput during training. The code of the CMER and the dataset will be released at \href{https://github.com/ZhangWeihang99/CMER}{[Link]}.
\end{abstract}




\maketitle

\section{Introduction}
\label{sec:introduction}
Continuous advances in earth observation technology and the increasing number of satellites have accelerated the acquisition of high-resolution, multi-source remote sensing (RS) image data \citep{chi2016big, amani2020google}. 
How to effectively extract critical information of interest from massive RS images has become an important research direction in RS image processing \citep{zhou2023remote, li2020hashing}.

RS image retrieval has been a hot research topic in the last decade  \citep{sun2022multisource, sun2023consistency, blumenstiel2024multi}. Many RS image retrieval methods have emerged, among which the cross-modal retrieval methods have gained growing interest from researchers due to their subjectivity. Concretely, the cross-modal retrieval methods contain multiple modalities in query, including multispectral images \citep{sun2024cross, xiong2022interpretable}, sketches \citep{fang2021cohesion}, audio \citep{mao2018deep, ning2021semantics}, ground panorama \citep{ren2023hashing, li2024unleashing}, and text \citep{liu2024text, ma2024direction}.

Remote sensing text--image retrieval (RSTIR) aims to retrieve RS images by corresponding descriptions \citep{zhang2023hypersphere}. According to the patterns of cross-modal interaction, RSTIR methods can be divided into two categories: dual-stream and single-stream methods. Text and images are encoded independently in the dual-stream methods \citep{zhang2023hypersphere, pan2023reducing}. The single-stream methods perform cross-modal feature fusion during the early feature extraction stage.
Recently, large pre-trained visual-language models have achieved significant success \citep{radford2021learning, li2021align, li2023blip}, bringing new insights for RSTIR \citep{li2024vision}. The parameter-efficient transfer learning (PETL) methods strive to transfer the prior information in pre-trained tasks to downstream tasks through fine-tuning a small set of parameters in the large pre-trained models \citep{lialin2023scaling}, which has gained widespread popularity. 

Although some progress has been made in RSTIR, one remaining challenge is improving the resource efficiency of transfer learning as the complexity of pre-trained models increases in RSTIR.

First, the current transfer learning methods for RSTIR suffer from suboptimal efficiency of GPU memory \citep{liu2024remoteclip, yuan2023parameter, djoufack2022clip}. Even if few parameters are fine-tuned in previous superior methods \citep{liu2024remoteclip, yuan2023parameter}, substantial intermediate activations still require computation and storage during training. In addition, the RS domain is more challenging during transfer learning. It is attributable to the high complexity of RS images, especially with multi-scale objects \citep{chen2023multiscale}. Conventional methods often encounter interference from background noise when extracting visual features of targets with a small pixel proportion. The Focus-Adapter is proposed to tackle the above issue. The Focus-Adapter adopts the side branch architecture to improve the efficiency of memory utilization. Moreover, the focus layer is devised to emphasize the regionally salient feature of small targets. It can suppress the interference of background pixels for small target feature extraction. 

To enhance data efficacy, we propose a plug-and-play augmentation technique that combines the specific content data with land cover categories. Recent mainstream RSTIR approaches neglect the substantial prior knowledge provided by the land cover categories \citep{pan2023reducing}. We regard RS scene categories as metadata tags \citep{espinoza2013earth} that serve as the prompt for the concrete query text. The scene label augmentation can shrink the retrieval space and mitigate mismatches between scenarios without increasing extra memory consumption. We also propose the negative sample recycling strategy to expand the negative sample pool. The sampling pool of negative samples in traditional methods is limited by the current mini-batch data, which is constrained by the available memory of hardware \citep{zhang2023hypersphere, yuan2023parameter}. We recycle the negative samples of previous iterations for the contrastive learning of current iteration data, which can better sample the underlying semantic space \citep{he2020momentum} of RS images.

Our contributions can be summarized as follows.
\begin{itemize}
    \item We propose a computation and memory-efficient retrieval framework for RSTIR. To the best of our knowledge, CMER is the first framework dedicated to improving resource efficiency beyond parameter efficiency during training.
    \item We propose the Focus-Adapter, a side branch architecture with the focus layer as the core component. The Focus-Adapter improves the efficiency of GPU memory and emphasizes the regionally salient feature of small targets.
    \item To enhance the efficacy of training data, we innovatively regard RS scene categories as metadata tags and recycle the negative samples of previous iterations.
    \item We extend the benchmark with some excellent methods and conduct extensive experiments, which show that our novel CMER framework outperforms recent methods by a clear margin.
\end{itemize}

\section{Related work}
\label{sec:relatedwork}
\subsection{Remote sensing text--image retrieval}
RSTIR aims to exploit the semantic information in textual data to retrieve the corresponding RS images. Currently, prevalent methods for RSTIR can be categorized into two types based on the patterns of cross-modal interaction: single-stream and dual-stream methods.

The dual-stream methods maintain independence in extracting image-text features without cross-modal feature fusion. Therefore, the dual-stream method can independently extract embedding vectors offline from different modalities.
A multi-language framework for RSTIR was proposed by \citet{al2022multilanguage}, which can support multi-language queries. To address the semantic noise in RSTIR, \citet{pan2023prior} designed a prior instruction representation framework that consists of two progressive attention encoder structures and a cluster-wise loss. \citet{zhang2023hypersphere} proposed a concise and effective alignment strategy based on curriculum learning. \citet{yuan2023parameter} investigated the PETL method for RSTIR and extended the benchmark.

The single-stream methods perform cross-modal feature fusion during the early feature extraction stage.  
\citet{yuan2022mcrn} designed a multi-source retrieval network that establishes the common space through multimodal shared coding. \citet{tang2023interacting} proposed the interacting-enhancing feature transformer for RSTIR. With increasing complexity, some recent RSTIR methods still have room for improvement in terms of efficiency and efficacy during training.
\subsection{Transfer learning}
The concept of transfer learning was first introduced by \citet{pratt1992discriminability}. Many transfer learning methods have emerged in the last decade. Among them, PETL methods have received increasing attention \citep{lei2024conditional}.

The PETL methods train a fraction of the model parameters, which may be a new set of introduced parameters or a subset of the original model parameters. 
\citet{houlsby2019parameter} proposed the Adapter modules in order to improve the efficiency of fine-tuning large pre-trained models in natural language processing. \citet{jiang2022cross} proposed the Cross-Modal Adapter for the multi-modal domain, which allows early interactions between multimodal. 
\citet{zaken2021bitfit} introduced a sparse-finetuning method named BitFit. 
\citet{hu2021lora} introduced the LoRA module for large language models, which reparameterizes the weight update matrix by the simple low-rank matrix. Similar to LoRA, \citet{edalati2022krona} proposed the Kronecker Adapter.
However, existing PETL algorithms are suboptimal in the more challenging RS domain because they are mainly tailored for downstream tasks, which are in the same domain as the pre-training tasks. 
\section{Method}
\begin{figure*}[!t]
    \centering
    \includegraphics [width=0.96\textwidth]{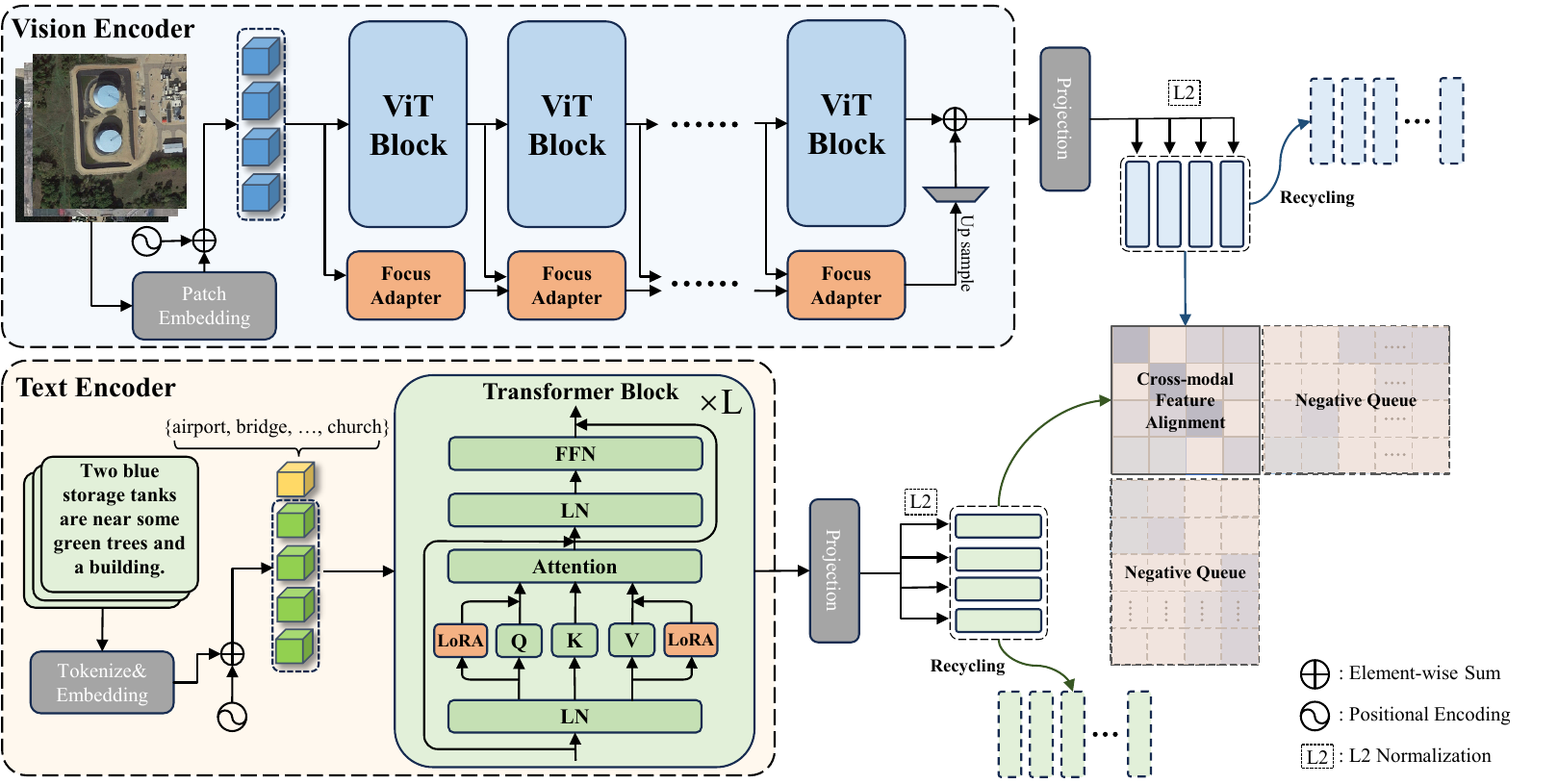}
    \caption{The pipeline of the proposed computation and memory-efficient retrieval (CMER) framework. 
    }
    \label{fig:pipline}
\end{figure*}

\label{sec:method}
\subsection{Formulation}
\label{sec:method_formulation}
Figure \ref{fig:pipline} illustrates the pipeline of our framework. For RSTIR, features of the input image $I$ and features of the description text $T$ are obtained after cross-modal feature extraction. The obtained semantic embedding $\boldsymbol{s}_e$ and visual embedding $\boldsymbol{v}_e$ are normalized to the unit hypersphere for similarity measurement. 
    \subsubsection{Visual representation}
    We utilize the Vision Transformer (ViT) \citep{dosovitskiy2020image} as our primary visual encoder, initialized with pre-trained CLIP weights. To extract visual embedding vectors, the RS images are initially partitioned into $N \times N$ patches. The initial visual embedding sequence $\boldsymbol{v}^0$ is obtained through sequence expansion and position encoding, which can be expressed as:
        \begin{equation}
            \label{equ:visual_init}
            \boldsymbol{v}^0 = [{\boldsymbol{i}_{class}};{\boldsymbol{W}_I}{\boldsymbol{i}_0}; \cdots ;{\boldsymbol{W}_I}{\boldsymbol{i}_{{N^2} - 1}}] + {\boldsymbol{i}_{pos}},
        \end{equation}
        where ${\boldsymbol{i}_{class}}$ is the added CLS token, ${\boldsymbol{i}_l}$ represents $l$-th image patch, and ${\boldsymbol{W}_I}$ is the image embedding matrix. $\boldsymbol{i}_{pos}$ is the position embedding vector. Subsequently, the initial visual embedding sequence ${\boldsymbol{v}^0}$ passes through ${D_v}$ blocks of the ViT, which can be represented as:
        \begin{equation}
            \label{equ:vit_msa}
            {{\tilde {\boldsymbol{v}}}^d} = { \operatorname{MSA}}({ \operatorname{LN}}({\boldsymbol{v}^{d - 1}})) + {\boldsymbol{v}^{d - 1}},d \in [1,{D_v}],
        \end{equation}
        \begin{equation}
            \label{equ:vit_ffn}
            {\boldsymbol{v}^d} = { \operatorname{FFN}}({ \operatorname{LN}}({{\tilde {\boldsymbol{v}}}^d})) + {{\tilde {\boldsymbol{v}}}^d},d \in [1,{D_v}].
        \end{equation}
        where $ \operatorname{MSA}(\cdot)$ represents the Multi-head Self Attention module \citep{vaswani2017attention}, $ \operatorname{LN}(\cdot)$ represents the layer normalization function and $ \operatorname{FFN}(\cdot)$ represents the Feed-Forwad Network. ${{\tilde {\boldsymbol{v}}}^d}$ indicates the intermediate hidden features extracted by MSA module in ${d}$-th ViT block. ${\boldsymbol{v}^d}$ is the final output of ${d}$-th block in ViT. The CLS token output by the $D_v$-th block is considered the image's global representation, and the visual embedding $\boldsymbol{v}_e$ is obtained after normalization and projection.

    \subsubsection{Text representation}
    BERT \citep{devlin2018bert} is utilized to obtain semantic embedding $\boldsymbol{s}_e$, which is initialized with pre-trained CLIP weights. The description text is tokenized based on byte pair encoding, which is prefixed with the BOS token and suffixed with the EOS token. After token embedding and position encoding, the initial description text tokens can be represented as: 
        \begin{equation}
            \label{equ:text_init}
            {\boldsymbol{s}^0} = [{{\boldsymbol{M}_E}\boldsymbol{t}_{bos}};{\boldsymbol{M}_E}{\boldsymbol{t}_0}; \cdots ;{\boldsymbol{M}_E}{\boldsymbol{t}_L};{\boldsymbol{M}_E}{\boldsymbol{t}_{eos}}] + {\boldsymbol{t}_{pos}},
        \end{equation}
        where $\boldsymbol{t}_{bos}$ refers to the token that represents the beginning of the sentence. ${\boldsymbol{t}_{eos}}$ refers to the token that represents the ending of the sentence. $\boldsymbol{t}_l$ stands for $l$-th token in the description text, and the preset maximum text processing length is denoted as $L$. $\boldsymbol{M}_E$ refers to the token embedding matrix. ${\boldsymbol{t}_{pos}}$ is the position embedding vector added to each token.

    The initial tokens sequence ${\boldsymbol{s}^0}$ is fed to transformer blocks to obtain semantic embedding, which can be denoted as:
        \begin{equation}
            \label{equ:trans_msa}
            {{\tilde {\boldsymbol{s}}}^d} = { \operatorname{MSA}}({ \operatorname{LN}}({\boldsymbol{s}^{d - 1}})) + {\boldsymbol{s}^{d - 1}},d \in [1,{D_s}],
        \end{equation}
        \begin{equation}
            \label{equ:trans_ffn}
            {\boldsymbol{s}^d} = { \operatorname{FFN}}({ \operatorname{LN}}({{\tilde {\boldsymbol{s}}}^d})) + {{\tilde {\boldsymbol{s}}}^d},d \in [1,{D_s}].
        \end{equation}

\subsection{Focus-Adapter}
\label{sec:method_focus_adapter}


Recent mainstream RSTIR methods based on transfer learning are sub-optimal regarding GPU memory efficiency during training \citep{yuan2023parameter}, \citep{djoufack2022clip}. Despite fine-tuning only a few parameters, a massive volume of intermediate activations still needs to be computed and stored during the training process. 

\begin{figure}[!t]
    \centering
    \includegraphics [width=0.40\linewidth]{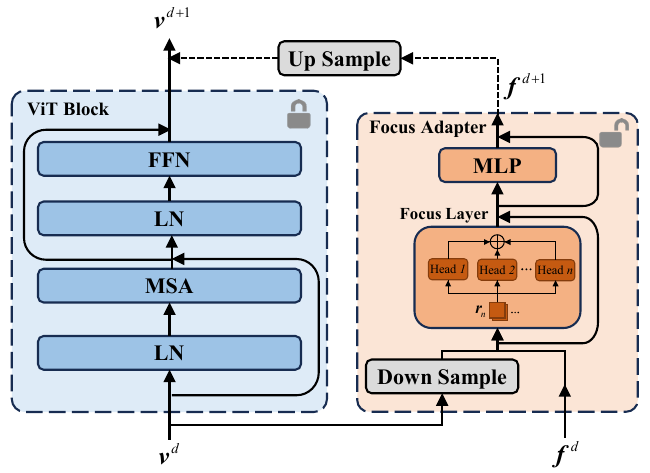}
    \caption{The implementation details of the proposed Focus-Adapter. 
    }
    \label{fig:focus_adapeter}
\end{figure}

For the sake of generality, we take a simple multi-layer perceptual network as an example and assume that layer $i$ is the linear layer with $\boldsymbol{z}_n = \boldsymbol{z}_{n-1}\boldsymbol{W}_n+b_n$. According to the chain rule, the variables required computation in the backward propagation process include:

    \begin{equation}
        \label{equ:grad_w}
        \frac{{\partial L}}{{\partial {\boldsymbol{W}_n}}} = \boldsymbol{z}_n^ \top \frac{{\partial L}}{{\partial {\boldsymbol{z}_{n + 1}}}},
    \end{equation}
      \begin{equation}
        \label{equ:grad_b}
        \frac{{\partial L}}{{\partial {b_n}}} = \frac{{\partial L}}{{\partial {\boldsymbol{z}_{n + 1}}}},
    \end{equation}
    \begin{equation}
        \label{equ:grad_a}
        \frac{{\partial L}}{{\partial {\boldsymbol{z}_n}}} = \frac{{\partial L}}{{\partial {\boldsymbol{z}_{n + 1}}}}\frac{{\partial {\boldsymbol{z}_{n + 1}}}}{{\partial {\boldsymbol{z}_n}}} = \frac{{\partial L}}{{\partial {\boldsymbol{z}_{n + 1}}}}\boldsymbol{W}_i^ \top.
    \end{equation}
    where $L$ represents the overall loss function, $\boldsymbol{W}_n$ and $b_n$ denote the weight matrix and bias vector of the $n$-th layer, respectively. $\boldsymbol{z}_n$ denotes the intermediate activation of the $n$th layer.

As demonstrated in Eq. (\ref{equ:grad_w}) and Eq. (\ref{equ:grad_a}), the memory footprint consists of two items: intermediate activation $\boldsymbol{z}_n$, and parameters $\boldsymbol{W}_n$ that need to be updated. Previous RSTIR methods \citep{yuan2023parameter} reduce the size of trainable parameters $\boldsymbol{W}_n$ but neglect the computation and storage of activation $\boldsymbol{z}_n$. 

To further enhance computation and memory efficiency, we propose the novel Focus-Adapter. As shown in Fig. \ref{fig:focus_adapeter}, inspired by \citet{zhang2020side}, deployment of the side branch decreases the computation and storage of activations when constructing the computational graph:

\begin{equation}
    \label{equ:focus_adapter_side_tuning}
    {\boldsymbol{h}^d} = {\boldsymbol{v}^d}{\boldsymbol{W}_{down}} + {\boldsymbol{f}^d} +  {b}, d \in [1,{D_f}].
\end{equation}
where $\boldsymbol{v}^d$ represents the feature of $d$-th block in the main branch calculated from Eq. (\ref{equ:vit_ffn}), and $\boldsymbol{f}^d$ represents the feature of $d$-th adapter in the side branch. ${\boldsymbol{W}_{down}}$ serves as the shared downsample matrix, and $b$ serves as the bias item.

\begin{figure}[!t]
    \centering
    \includegraphics [width=0.40\linewidth]{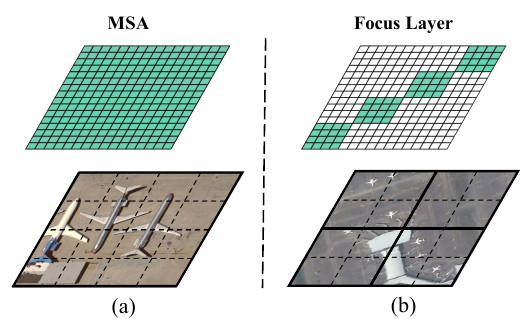}
    \caption{Region attention mechanism in the focus layer. The dotted lines in the RS images indicate the division of the image patches. The solid rectangles in the RS images represent the range of attention modeling. The above grid diagram shows the modeling relationships between image patches.}
    \label{fig:focus_layer}
\end{figure}
Moreover, multi-scale variation is one of the main properties in RS images \citep{chen2023multiscale}. For images with relatively large regions of interest, as shown in Fig. \ref{fig:focus_layer}(a), it seems reasonable to utilize features of all patches to compute the attention score when extracting visual features. However, for images with smaller and more dispersed targets, as shown in Fig. \ref{fig:focus_layer}(b), long-range modeling \citep{vaswani2017attention} may introduce background interference and computational redundancy. 

Inspired by the local inductive bias of the CNN \citep{he2016deep}, the focus layer is designed to reduce the interference from the background by selectively emphasizing the regionally salient feature of small targets.
Specifically, we introduce the local inductive bias of CNN \citep{he2016deep}. The feature $\boldsymbol{h}^d$ calculated from Eq. (\ref{equ:focus_adapter_side_tuning}) is first partitioned into distinct regions $\left[\boldsymbol{r}_1, \boldsymbol{r}_2, \cdots, \boldsymbol{r}_N \right]$ according to the size of focus field. As shown in Fig. \ref{fig:focus_layer}, the representation of regionally salient features is enhanced by the region attention mechanism, which is aimed at suppressing background pixel interference in long-range modeling. The above process can be represented as:
\begin{equation}
    \label{equ:mh_mlp}
    {head}_i = \left[{\boldsymbol{r}_1},{\boldsymbol{r}_2},\cdots,{\boldsymbol{r}_N} \right]{\boldsymbol{W}_i} + {b_i},i \in [0,H],
\end{equation}
\begin{equation}
    \label{equ:focus_adapter_lmsa}
    {{\tilde {\boldsymbol{f}}}^d} = \left[{head}_0,{head}_1,\cdots,{head}_H\right]+\boldsymbol{h}^{d - 1},d \in [1,{D_f}],
\end{equation}
\begin{equation}
    \label{equ:focus_adapter_ffn}
    {\boldsymbol{f}^d} = {{\tilde {\boldsymbol{f}}}^d}{\boldsymbol{W}_d}+ {{\tilde {\boldsymbol{f}}}^d},d \in [1,{D_f}].
\end{equation}
where ${head}_i$ represents the salient feature emphasized by $i$-th attention head, and the enhanced region feature in $d$-th Focus-Adapter is denoted as ${{\tilde {\boldsymbol{f}}}^d}$. The total number of the Focus-Adapter is ${D_f}$. 

\subsection{Scene label augmentation}
\label{sec:method_label_prompt}
Recent mainstream RSTIR approaches primarily concentrate on specific content \citep{sudha2019review}, neglecting the land cover categories. Land cover categories can provide substantial prior knowledge and play an important role in RS image retrieval \citep{cheng2017remote}. 
To shrink the search space and alleviate the difficulties in aligning cross-modal features, we propose scene label augmentation, which aims to provide prior knowledge of scene categories. 

More specifically, we regard RS scene categories as metadata tags and combine the metadata with fine-grained content data. Different scenario categories are added in front of content-based descriptions as the prompt \citep{yuan2023parameter}, as shown in Fig. \ref{fig:pipline}. 
The augmented captions include scene categories and content information that can be represented as:
\begin{equation}
    \label{equ:text_concat}
    {\boldsymbol{t}} = {\operatorname{Concat}({\boldsymbol{t}_{meta}},{\boldsymbol{t}_{cont}})},
\end{equation}
\begin{equation}
    \label{equ:label_prompt}
    {\boldsymbol{s}^0} = {\boldsymbol{M}_E}\cdot{\boldsymbol{t}} + {\boldsymbol{T}_{pos}}.
\end{equation}
where $\boldsymbol{t}_{meta}$ refers the prompt token of metadata, $\boldsymbol{t}_{cont}$ represents the token of fine-grained content data, $\boldsymbol{M}_E$ refers the word embedding matrix of BERT, $\boldsymbol{T}_{pos}$ is the position embedding matrix and $\boldsymbol{s}^0$ represents the semantic feature after applying scene label augmentation.

It is worth noting that the designed prompt of the scenario category does not require training. Its vector representation is embedded by the word embedding matrix, which does not require computing gradients and updates. Therefore, our concise and effective augmentation method not only does not incur additional computational overhead but also leverages the prior knowledge contained in scene categories.
\begin{figure}[!t]
    \centering
    \includegraphics [width=0.40\linewidth]{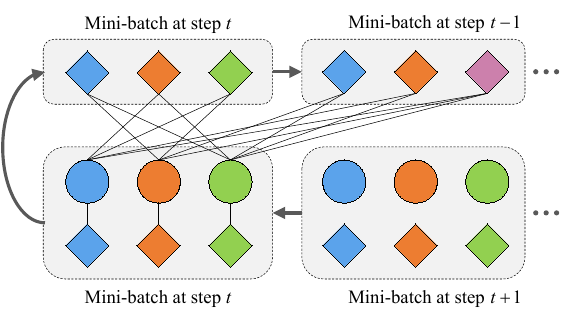}
    \caption{The queue update process in the negative sample recycling strategy. 
    Diamonds represent visual features, and semantic features are represented by circles. The same color indicates the same category. 
    }
    \label{fig:neg_queue}
\end{figure}
\subsection{Negative sample recycling}
In traditional methods, the sampling pool of negative samples is limited to the size of mini-batch data, which is constrained by the available GPU memory \citep{pan2023reducing}. 
To address this issue, inspired by \citep{he2020momentum}, we innovatively propose the negative sample recycling strategy to expand the pool of negative samples. 
We establish two distinct negative queues $\boldsymbol{Q}_v$ and $\boldsymbol{Q}_s$ to temporarily preserve the negative samples for visual and semantic embeddings, respectively. As shown in Fig. \ref{fig:neg_queue}, the training data at the previous time step $t-1$ is recycled as negative samples for the contrastive learning of current data at step $t$. The queues $\boldsymbol{Q}_v$ and $\boldsymbol{Q}_s$ are updated once per iteration, and negative samples from different iterations are maintained first in, first out. The proposed two negative queues $\boldsymbol{Q}_v$ and $\boldsymbol{Q}_s$ expand the negative sample pool by decoupling it from the mini-batch size. It is achieved without introducing additional encoders and considerable computational loads. 

For one positive pair $(\boldsymbol{s}_e, \boldsymbol{v}_e)$ in text--image retrieval, the negative sample pool is defined as the samples from the negative queue $\boldsymbol{Q}_v$ that have different scenario categories compared to the positive sample $\boldsymbol{v}_e$. For training stability, we adopt the difficulty-weighted hinge loss \citep{zhang2023hypersphere} to measure the alignment between the current samples and the negative samples from queues $\boldsymbol{Q}_v$ and $\boldsymbol{Q}_s$:
\begin{equation}
    \label{equ:dwh_t2i}
    {l_v} = {[\alpha  - S(\boldsymbol{s}_e,\boldsymbol{v}_e) + S(\boldsymbol{s}_e,{\boldsymbol{q}_v})]_ + },
\end{equation}
\begin{equation}
    \label{equ:dwh_i2t}
    {l_s} = {[\alpha  - S(\boldsymbol{v}_e,\boldsymbol{s}_e) + S(\boldsymbol{v}_e,{\boldsymbol{q}_s})]_ + },
\end{equation}
\begin{equation}
    \label{equ:dwh}
    {L_{queue}} = \sum\limits_{{\boldsymbol{q}_v} \in \boldsymbol{Q}_v} {{l_v}{e^{ - \beta {l_v}}}}  + \sum\limits_{\boldsymbol{q}_s \in \boldsymbol{Q}_s} {{l_s}{e^{ - \beta {l_s}}}}.
\end{equation}
where $(\boldsymbol{s}_e,\boldsymbol{v}_e)$ is one positive sample pair, $(\boldsymbol{s}_e,{\boldsymbol{q}_v})$ is one negative sample pair, $\boldsymbol{q}_v$ and $\boldsymbol{q}_s$ are negative samples in the queues $\boldsymbol{Q}_v$ and $\boldsymbol{Q}_s$, and margin $\alpha$ controls the margin of the feature alignment. $S(\cdot)$ stands for cosine similarity function. ${[x]_ + } \equiv \max (x,0)$.

The $L_{batch}$ loss measures the alignment error when the negative samples are from the current batch data, which can be expressed as: 
\begin{equation}
    \label{equ:l_infonce_t2i}
    {p_{t2i}}(I,T) = \frac{{\exp (S(I,T)/\tau )}}{{\exp (S(I,T)/\tau ) + \sum\limits_{{\hat I}} {\exp (S({\hat I},T)/\tau )} }},
\end{equation}
\begin{equation}
    \label{equ:l_infonce_i2t}
    {p_{i2t}}(I,T) = \frac{{\exp (S(I,T)/\tau )}}{{\exp (S(I,T)/\tau ) + \sum\limits_{{\hat T}} {\exp (S(I,{\hat T})/\tau )} }},
\end{equation}
\begin{multline}
    \label{equ:l_infonce_total}
    {L_{batch}} =   - \frac{1}{2}{E_{(I,T) \sim D}}[ \log ({p_{i2t}}(I,T)) \\
     + \log ({p_{t2i}}(I,T))].
\end{multline}
where $(I,T)$ refers the positive pair, $\hat I$ and $\hat T$ refers the negative sample from current batch, and $\tau$ servers as the temperature coefficient. ${p_{t2i}}(I,T)$ represents the matching probability of the positive pair $(I, T)$ when performing text-image retrieval. 

The negative sample queues $\boldsymbol{Q}_v$ and $\boldsymbol{Q}_s$ are updated by current batch data. The parameters $\theta_f$ in the Focus-Adapter and the parameters $\theta_l$ in the LoRA module are updated according to the final objective function $L$:
\begin{equation}
    \label{equ:final_loss}
    L = {L_{batch}} + {L_{queue}}.
\end{equation}

\section{Experiments and analysis}
\label{sec:experiment}
\label{sec:exp_comparison}

\subsection{Datasets and protocols}
\label{sec:exp_dataset}
Our proposed model is validated on two standard RSTIR datasets, RSICD and RSITMD.
The RSICD, proposed by \citet{lu2017exploring}, comprises 10,921 images. Each image in the RSICD is associated with five descriptive texts, yielding $10,921 \times 5$ paired samples. 
Compared to the RSICD, the RSITMD \citep{yuan2021exploring} is smaller and more fine-grained in query.
The RSITMD consists of 4,743 images. Each image in the RSITMD is associated with five descriptive texts, yielding $4,743 \times 5$ paired samples. The resolution of images in the RSITMD is standardized to $256 \times 256$ pixels.

To reflect the training consumption, the number of training parameters (M), peak GPU memory usage (MB), and the throughput of training data are introduced (pairs/s). The evaluation metrics for model performance are the widely recognized $R@k$ and $mR$:
\begin{equation}
    \label{equ:mr}
    mR = \frac{1}{\left |K \right |}\sum\limits_{k \in K} {R@k}.
\end{equation}
where $k$ usually takes the common values of 1, 5, and 10. $mR$ denotes the average value for different values $R@k$ where $k$ belongs to the common value set $K$.
\subsection{Implementation details}
All experiments in this paper are implemented on two NVIDIA V100 GPUs, and the weights that achieve the best performance on the validation set are saved. Taking our proposed CMER with the ViT-B-16 backbone as an example, we fix three random seeds to ensure repeatability and reliability. The average of three runs with different random seeds represents the overall performance. 
We choose AdamW as the optimizer and train with a batch size of 256 for 20 epochs. The initial learning rate is $5 \times 10^{-4}$.
\begin{table*}[!t]
  \centering
  \caption{Performance of Different Methods on the RSITMD Test Set}
  \resizebox{\textwidth}{!}{
    \begin{tabular}{lccccccccc}
    \toprule
        \multicolumn{1}{l}{\multirow{2}[2]{*}{\textbf{Methods}}} & \multirow{2}[2]{*}{\textbf{Backbone (Vision / Text)}} & \multirow{2}[2]{*}{\textbf{ Training Params}} & \multicolumn{3}{c}{\textbf{Image-Query-Text}} & \multicolumn{3}{c}{\textbf{Text-Query-Image}} & \multirow{2}[2]{*}{\textbf{mR}} \\
          &       &       & \textbf{R@1} & \textbf{R@5} & \textbf{R@10} & \textbf{R@1} & \textbf{R@5} & \textbf{R@10} &  \\
    \midrule
    \multicolumn{10}{c}{\textbf{CNN-based methods}} \\
    \midrule
    AMFMN  & ResNet-18 / GRU & 36.70M & 10.63  & 24.78  & 41.81  & 11.51  & 34.69  & 54.87  & 29.72  \\
    MCRN   & ResNet-18 / GRU & 52.35M & 13.27  & 29.42  & 41.59  & 9.42  & 35.53  & 52.74  & 30.33  \\
    HVSA   & ResNet-18 / GRU & 35.01M & 13.20  & 32.08  & 45.58  & 11.43  & 39.20  & 57.45  & 33.16  \\
    SWAN  & ResNet-50 / GRU &   -    & 13.35  & 32.15  & 46.90  & 11.24  & 40.40  & 60.60  & 34.11  \\
    \midrule
    \multicolumn{10}{c}{\textbf{Transformer-based methods}} \\
    \midrule
    Single Language   &  ViT-B-32 / BERT & 151M  & 19.69  & 40.26  & 54.42  & 17.61  & 49.73  & 66.59  & 41.38  \\
    IEFT   & Vision Trans / BERT  & 100.12M & 11.19  & 38.09  & 58.84  & 15.49  & 37.61  & 51.40  & 35.44  \\
    PIR    & Swin-T+ResNet50 / BERT &   -    & 18.14  & 41.15  & 52.88  & 12.17  & 41.68  & 63.41  & 38.24  \\
    Linear probe  &  ViT-B-32 / BERT & 0.53M & 13.71  & 33.41  & 48.01  & 10.97  & 36.85  & 56.15  & 33.18  \\
    Cross-Modal Adapter   &  ViT-B-32 / BERT & 0.16M & 18.16  & 36.08  & 48.72  & 16.31  & 44.33  & 64.75  & 38.06  \\
    PE-RSITR  &  ViT-B-32 / BERT & 0.16M & 23.67  & 44.07  & 60.36  & 20.10  & 50.63  & 67.97  & 44.47  \\
    CLIP-adapter   &  ViT-B-32 / BERT & 3.04M & 20.57  & 42.03  & 54.86  & 17.78  & 46.68  & 62.56  & 40.75  \\
    CLIP-lora   &  ViT-B-32 / BERT & 3.93M & 21.68  & 41.37  & 56.19  & 16.85  & 47.25  & 67.74  & 41.85  \\
    CLIP-bitfit  &  ViT-B-32 / BERT & 0.17M & 21.01  & 42.25  & 52.21  & 14.95  & 45.13  & 64.29  & 39.97  \\
    CLIP-adapter   &  ViT-B-16 / BERT & 3.04M & 21.90  & 41.81  & 53.09  & 17.83  & 46.94  & 66.19  & 41.29  \\
    CLIP-lora   &  ViT-B-16 / BERT & 3.93M & 22.34  & 43.14  & 57.74  & 19.02  & 51.46  & 69.69  & 43.90  \\
    CLIP-bitfit  &  ViT-B-16 / BERT & 0.17M & 23.67  & 42.10  & 55.45  & 18.33  & 48.12  & 67.35  & 42.50  \\
    CLIP-adapter   &  ViT-L-14 / BERT & 6.82M & 25.44  & 41.15  & 54.64  & 18.80  & 50.08  & 67.61  & 42.95  \\
    CLIP-lora  &  ViT-L-14 / BERT & 8.65M & 26.99  & 45.35  & 56.63  & 21.94  & 55.26  & 71.72  & 46.32  \\
    CLIP-bitfit  &  ViT-L-14 / BERT & 0.37M & 26.54  & 41.15  & 54.64  & 20.04  & 51.15  & 69.29  & 43.80  \\
    \midrule
    CMER(ours) &  ViT-B-32 / BERT & 2.72M & 21.82  & 47.64  & 62.24  & 16.38  & 53.49  & 79.70  & 46.88  \\
    CMER(ours) &  ViT-B-16 / BERT & 2.72M & 21.90  & 48.89  & 62.46  & 18.30  & 55.36  & 80.82  & 47.96  \\
    CMER(ours) &  ViT-L-14 / BERT & 4.50M & 23.67  & 49.77  & 63.93  & 17.25  & 56.41  & \textbf{81.15} & 48.70  \\
    CMER(ours) &  ViT-H-14 / BERT & 9.14M & \textbf{27.87} & \textbf{54.64} & \textbf{65.92} & \textbf{23.14} & \textbf{58.45} & 80.75  & \textbf{51.80} \\
    \bottomrule
    \end{tabular}}%
  \label{tab:exp_rsitmd}%
\end{table*}%
\subsection{Performance comparisons}
\begin{table*}[!t]
  \centering
  \caption{Performance of Different Methods on the RSICD Test Set}
  \resizebox{\textwidth}{!}{
    \begin{tabular}{lccccccccc}
    \toprule
    \multicolumn{1}{l}{\multirow{2}[2]{*}{\textbf{Methods}}} & \multirow{2}[2]{*}{\textbf{Backbone (Vision / Text)}} & \multirow{2}[2]{*}{\textbf{ Training Params}} & \multicolumn{3}{c}{\textbf{Image-Query-Text}} & \multicolumn{3}{c}{\textbf{Text-Query-Image}} & \multirow{2}[2]{*}{\textbf{mR}} \\
          &       &       & \textbf{R@1} & \textbf{R@5} & \textbf{R@10} & \textbf{R@1} & \textbf{R@5} & \textbf{R@10} &  \\
    \midrule
    \multicolumn{10}{c}{\textbf{CNN-based methods}} \\
    \midrule
    AMFMN & ResNet-18 / GRU & 36.70M & 5.21  & 14.72  & 21.57  & 4.08  & 17.00  & 30.60  & 15.53  \\
    MCRN   & ResNet-18 / GRU & 52.35M & 6.59  & 19.40  & 30.28  & 5.03  & 19.38  & 32.99  & 18.95  \\
    HVSA   & ResNet-18 / GRU & 35.01M & 7.47  & 20.62  & 32.11  & 5.51  & 21.13  & 34.13  & 20.16  \\
    SWAN   & ResNet-50 / GRU &   -    & 7.41  & 20.13  & 30.86  & 5.56  & 22.26  & 37.41  & 20.61  \\
    \midrule
    \multicolumn{10}{c}{\textbf{Transformer-based methods}} \\
    \midrule
    Single Language  &  ViT-B-32 / BERT & 151M  & 10.70  & 29.64  & 41.53  & 9.14  & 28.96  & 44.59  & 27.42  \\
    IEFT  & Vision Trans / BERT  & 100.12M & 8.38  & 28.17  & 44.16  & 8.78  & 28.47  & 43.88  & 26.97  \\
    PIR   & Swin-T+ResNet50 / BERT &   -    & 9.88  & 27.26  & 39.16  & 6.97  & 24.56  & 38.92  & 24.46  \\
    Linear probe  &  ViT-B-32 / BERT & 0.53M & 8.46  & 24.41  & 37.72  & 7.81  & 25.89  & 42.47  & 24.46  \\
    Cross-Modal Adapter  &  ViT-B-32 / BERT & 0.16M & 11.18  & 27.31  & 40.62  & 9.57  & 30.74  & 48.36  & 27.96  \\
    PE-RSITR &  ViT-B-32 / BERT & 0.16M & 14.13  & 31.51  & 44.78  & 11.63  & 33.92  & 50.73  & 31.12  \\
    CLIP-adapter  &  ViT-B-32 / BERT & 3.04M & 13.99  & 31.01  & 44.46  & 10.21  & 31.54  & 46.62  & 29.64  \\
    CLIP-lora   &  ViT-B-32 / BERT & 3.93M & 14.82  & 32.66  & 44.92  & 12.04  & 33.87  & 49.75  & 31.34  \\
    CLIP-bitfit  &  ViT-B-32 / BERT & 0.17M & 13.72  & 32.47  & 44.83  & 11.45  & 32.46  & 49.02  & 30.66  \\
    CLIP-adapter  &  ViT-B-16 / BERT & 3.04M & 14.36  & 31.65  & 44.46  & 11.60  & 32.68  & 48.32  & 30.51  \\
    CLIP-lora &  ViT-B-16 / BERT & 3.93M & 16.01  & 33.66  & 44.83  & 11.65  & 33.52  & 49.64  & 31.55  \\
    CLIP-bitfit &  ViT-B-16 / BERT & 0.17M & 15.55  & 33.02  & 47.30  & 11.60  & 33.79  & 49.97  & 31.87  \\
    CLIP-adapter  &  ViT-L-14 / BERT & 6.82M & 15.00  & 34.03  & 47.39  & 11.83  & 34.78  & 51.01  & 32.34  \\
    CLIP-lora   &  ViT-L-14 / BERT & 8.65M & 15.73  & 35.95 & 48.58  & \textbf{13.10}  & \textbf{38.40} & 53.37  & 34.19  \\
    CLIP-bitfit &  ViT-L-14 / BERT & 0.37M & 16.10  & \textbf{36.13}  & 48.85  & 12.64  & 36.26  & 53.04  & 33.84  \\
    \midrule
    CMER(ours) &  ViT-B-32 / BERT & 2.72M & 13.29  & 31.01  & 43.15  & 9.76  & 32.58  & 50.81  & 30.10  \\
    CMER(ours) &  ViT-B-16 / BERT & 2.72M & 14.15  & 32.29  & 46.56  & 10.61  & 32.10  & 50.33  & 31.01  \\
    CMER(ours) &  ViT-L-14 / BERT & 4.50M & 15.55  & 35.49  & \textbf{49.03} & 10.66  & 33.41  & 51.49  & 32.61  \\
    CMER(ours) &  ViT-H-14 / BERT & 9.14M & \textbf{17.56} & 35.22  & 48.67  & 12.66 & 37.54  & \textbf{55.24} & \textbf{34.48} \\
    \bottomrule
    \end{tabular}}%
  \label{tab:comp_rsicd}%
\end{table*}%

The proposed CMER is compared with recent well-performed RSTIR methods in this subsection. The results are shown in Tables \ref{tab:exp_rsitmd} and \ref{tab:comp_rsicd}. 

\textbf{Results on RSITMD:} In Table \ref{tab:exp_rsitmd}, we show the results of CMER as well as comparisons with the recent superior methods on the RSITMD test set. 
The best results are highlighted in bold.
From the results, our proposed CMER surpasses the recent state-of-the-art method with a clear margin. 
When compared with CNN-based methods, our CMER method outperforms SWAN by 12.77\% in overall retrieval performance (mR). 
When compared with Transformer-based methods on ViT-B-32, our CMER method outperforms the Single Language method by 13.11\% in text-to-image retrieval ({R@10}) and 5.50\% in overall retrieval performance. The proposed CMER (ViT-B-32) also surpasses the previous state-of-the-art method PE-RSITR by 11.73\% in text-to-image retrieval ({R@10}) and 2.41\% in overall retrieval performance. 
Our CMER achieves the best mR of 51.80\% on ViT-H-14. 

\begin{table*}[htbp]
    \centering
    \caption{Efficiency Of Different Methods}
    \resizebox{0.96\linewidth}{!}{
      \begin{tabular}{l|ccccccccc}
      \toprule
            & CLIP-adapter & CLIP-bitfit & CLIP-lora & CLIP-adapter & CLIP-bitfit & CLIP-lora & CMER(ours) & CMER(ours) & CMER(ours) \\
      \midrule
      Vison Backbone & ViT-B-16 & ViT-B-16 & ViT-B-16 & ViT-L-14 & ViT-L-14 & ViT-L-14 & ViT-B-16 & ViT-L-14 & ViT-H-14 \\
      \midrule
      Training Params(M) & 3.04  & \textbf{0.17}  & 3.93  & 6.82  & 0.37  & 8.65  & 2.72  & 4.50  & 9.14  \\
      \midrule
      Memory(MB) & 6841  & 6921  & 7173  & 22015  & 21740  & 22435  & \textbf{3488}  & 12450  & 18235  \\
      \midrule
      Throughput(pairs/s) & 200   & 191   & 137   & 48    & 47    & 35    & \textbf{276}   & 83    & 35  \\
      \bottomrule
      \end{tabular}}
    \label{tab:consume}
  \end{table*}%
  

\textbf{Results on RSICD:} We report results for the RSICD test set in Table \ref{tab:comp_rsicd}, where the proposed CMER outperforms many recent advanced methods. Our method is remarkably ahead of traditional CNN-based methods. When compared with Transformer-based methods, our CMER (ViT-B-32) outperforms the IEFT by 3.13\% in metric mR. Compared to CLIP-adapter, our CMER (ViT-B-32) achieves a relative improvement of 8.98\% in image retrieval (R@10).
Our CMER (ViT-H-14) achieves a 3.50\% relative improvement in text-to-image retrieval (R@10) compared to the recent excellent model CLIP-lora (ViT-L-14). 
\textbf{Comparison of Efficiency:} Table \ref{tab:consume} shows the efficiency of different methods during the training process. 
Compared with the excellent method CLIP-lora on ViT-B-16, our proposed CMER method reduces memory usage by 51\% and has a 2.0 times data throughput. Our proposed CMER also has a 1.4 times data throughput and reduces memory usage by 49\% compared to the baseline method CLIP-adapter on the ViT-B-16. The training resource consumption of CMER on ViT-H is on the same level as that of traditional methods on ViT-L. 
Our CMER is not only more efficient in resources but also has a competitive overall performance. Combined with the results in Tables \ref{tab:exp_rsitmd} and \ref{tab:comp_rsicd}, our approach achieves competitive results under the same constraints of resource consumption, realizing a better trade-off between resource consumption and performance.


\subsection{Ablation studies}
\label{subs:abla}
\begin{table*}[!t]
  \centering
  \caption{Ablation Experiments of CMER on the RSICD and RSITMD}
  \resizebox{\textwidth}{!}{
    \begin{tabular}{c|c|c|cc|cc|cc}
    \toprule
    \multirow{2}[2]{*}{Ablation Model} & Semantic Encoder & Visual Encoder & \multicolumn{2}{c|}{Training Data} & \multicolumn{2}{c|}{Consumption} & \multicolumn{2}{c}{mR} \\
     & LoRA  & Focus-Adapter & Negative Recycling & Scene Prompt  & Memory & Throughout & RSICD & RSITMD \\
    \midrule
    v1 & \usym{1F5F8}   &  &       &       & 13153 MB & 338 pairs/s  & 25.94 & 34.07  \\
    v2 & \usym{1F5F8}   & \usym{1F5F8} &       &       & 19327 MB& 276 pairs/s   & 29.49 & 40.59  \\
    d1 & \usym{1F5F8}   & \usym{1F5F8} & \usym{1F5F8}   &       & 19331 MB & 274 pairs/s   & 29.76 & 40.81  \\
    d2 & \usym{1F5F8}   & \usym{1F5F8} & \usym{1F5F8}   & \usym{1F5F8}   & 19331 MB & 274 pairs/s   & \textbf{31.01} & \textbf{47.96} \\
    \bottomrule
    \end{tabular}}%
  \label{tab:ablations}%
\end{table*}%
\subsubsection{Analyze Each Component in the CMER}
To systematically explore the effect of different components in the CMER, we have performed ablation experiments on both the RSICD and RSITMD, as shown in Table \ref{tab:ablations}. 


The semantic encoder remains consistent across all experiments, consisting of the LoRA modules and the BERT model \citep{hu2021lora}.
Comparing the v1 model with the v2, the introduction of the Focus-Adapter in v2 improves the mR metric by 3.55\% on the RSICD and 6.52\% on the RSITMD. 
The proposed Focus-Adapter effectively provides the complementary region information to the backbone and suppresses background interference.

The comparison between the v2 and the d1 models demonstrates the validity of the negative sample recycling strategy.
During training, the peak GPU memory usage increases by only 4MB, and the data throughput decreases by 2 pairs/s. The proposed negative sample recycling strategy decouples the sampling pool of negative samples from the batch size, which contributes to sampling the high-dimensional semantic space.

Compared to the d2 model with the d1, the scene label augmentation has brought improvements of 1.25\% and 7.15\% in the mR metric for the RSICD and RSITMD, respectively. 
Notably, the d2 and d1 models have the same memory footprint and data throughput. 
Our plug-and-play augmentation technique effectively shrinks the search space, significantly improves the retrieval accuracy, and does not involve additional computational overhead. 


\subsubsection{Hyperparameters of the Focus-Adapter}
Different hidden layer dimensions and sizes of the focus field in the focus layer have varying impacts on the CMER performance, as shown in Table \ref{tab:hp_focus_adapter}. The pre-trained ViT-B-16 model is utilized on the RSITMD to perform a preliminary grid search for hyperparameters of the Focus-Adapter. The TQI in Table \ref{tab:hp_focus_adapter} represents the performance of Text-Query-Image. We sum all R@k in text-query-image to obtain the TQI, and the IQT is calculated similarly. The hidden dimension takes common values of 128, 192, and 256, corresponding to the number of heads in multi-head attention as 4, 6, and 8, respectively. The head's dimension is fixed at 32. The size of focus fields are set to 1, 2, and 7. When the dimension of the hidden layer is 192, and the size of the focus field is 2, the best performance is obtained.
\begin{table}[!t]
  \centering
  \caption{Experiments of the Focus Layer.}
    \begin{tabular}{ccc|ccc}
    \toprule
    Hidden Dim & Focus Field & Head & IQT   & TQI   & mR \\
    \midrule
    128   & 1     & 4     & 113.69 & 127.46 & 40.19 \\
    128   & 2     & 4     & 113.34 & 128.36 & 40.28 \\
    128   & 7     & 4     & 112.51 & 127.21 & 39.95 \\
    192   & 1     & 6     & 114.75 & 127.38 & 40.35 \\
    192   & 2     & 6     & 114.06 & \textbf{129.49} & \textbf{40.59} \\
    192   & 7     & 6     & 112.08 & 127.46 & 39.92 \\
    256   & 1     & 8     & 113.48 & 127.79 & 40.21 \\
    256   & 2     & 8     & 110.82 & 129.32 & 40.02 \\
    256   & 7     & 8     & \textbf{114.89} & 126.45 & 40.22 \\
    \bottomrule
    \end{tabular}
  \label{tab:hp_focus_adapter}%
\end{table}%
\begin{table}[!t]
  \centering
  \caption{Grid Search for the Length of Queues on the RSITMD}
    \begin{tabular}{c|cccc}
    \toprule
    Length  & 0     & 2     & 4     & 8 \\
    \midrule
    IQT   & \textbf{114.06} & 113.43 & 113.51 & 113.91 \\
    \midrule
    TQI   & 129.49 & 130.66 & \textbf{131.36} & 130.48 \\
    \midrule
    mR    & 40.59 & 40.68 & \textbf{40.81} & 40.73 \\
    \bottomrule
    \end{tabular}%
  \label{tab:hp_negtive_queue}%
\end{table}%
\subsubsection{Hyperparameters of the negative queues}
We explore the impact of queue length in the proposed negative sample recycling strategy, and the results on the RSITMD are listed in Table \ref{tab:hp_negtive_queue}.
The appropriate length for the negative sample queue should be 4 times the batch size, which is the sum of the negative samples over 4 iterations. When the negative sample pool is expanded to four times the batch size, the text-to-image retrieval performance increases by 1.87 points, and the overall performance increases by 0.22\%. However, when too many negative samples are recycled, it leads to performance degradation instead. We suggest that excessively stale data hinders the learning of discriminative features from the currently useful data as parameters are updated. 

\subsection{Qualitative analysis}
\subsubsection{Focus-Adapter}
To reflect the contribution of the Focus-Adapter intuitively, we visualize the visual features extracted by the visual encoder. For each text query, heatmaps of the different methods on the positive pairs are visualized by the Grad-Cam \citep{selvaraju2017grad}.

\begin{figure*}[!t]
    \centering
    \includegraphics [width=0.95\textwidth]{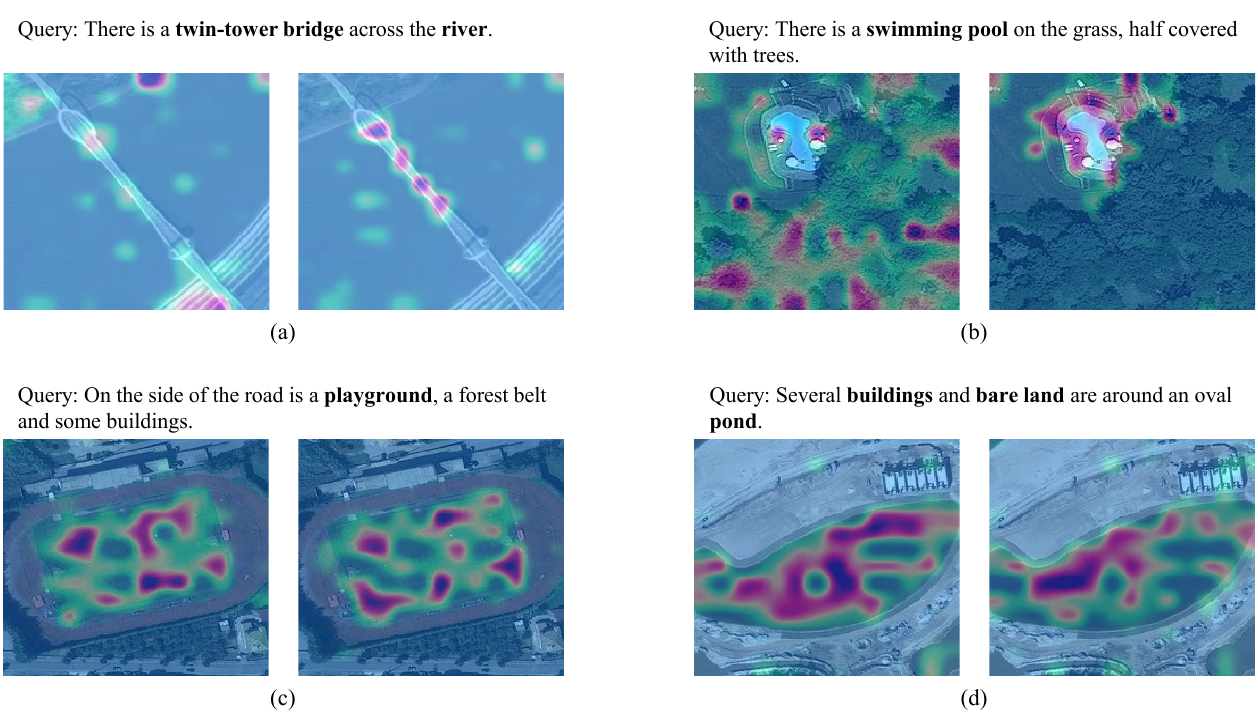}
    \caption{Qualitative results of the proposed Focus-Adapter. 
    In (a) and (b), the primary entities described in the text are relatively small in scale, with a large proportion of background pixels. In (c) and (d), the primary entities described in the text are relatively large in scale.}
    \label{fig:grad-cam}
\end{figure*}

In Fig. \ref{fig:grad-cam}(a), compared to the v1 method in subsection \ref{subs:abla}, our approach effectively filters the interference of river pixels, focusing more on the bridge. Similarly to Fig. \ref{fig:grad-cam}(a), in Fig. \ref{fig:grad-cam}(b), the baseline method does not focus well on the swimming pool, mainly due to interference from trees and grass, while our proposed method performs better. As shown in Fig. \ref{fig:grad-cam}(c) and Fig. \ref{fig:grad-cam}(d), the primary entities are the football field and pond, which have a relatively large pixel proportion.
Our method also performs well in RS images with relatively large regions of interest. 
The above visualization experiments further demonstrate the effectiveness of the focus layer in suppressing background interference when extracting features of small entities.

\subsubsection{Modal entanglement visualization}
\begin{figure}[!t]
  \centering
  \includegraphics[width=0.23\textwidth]{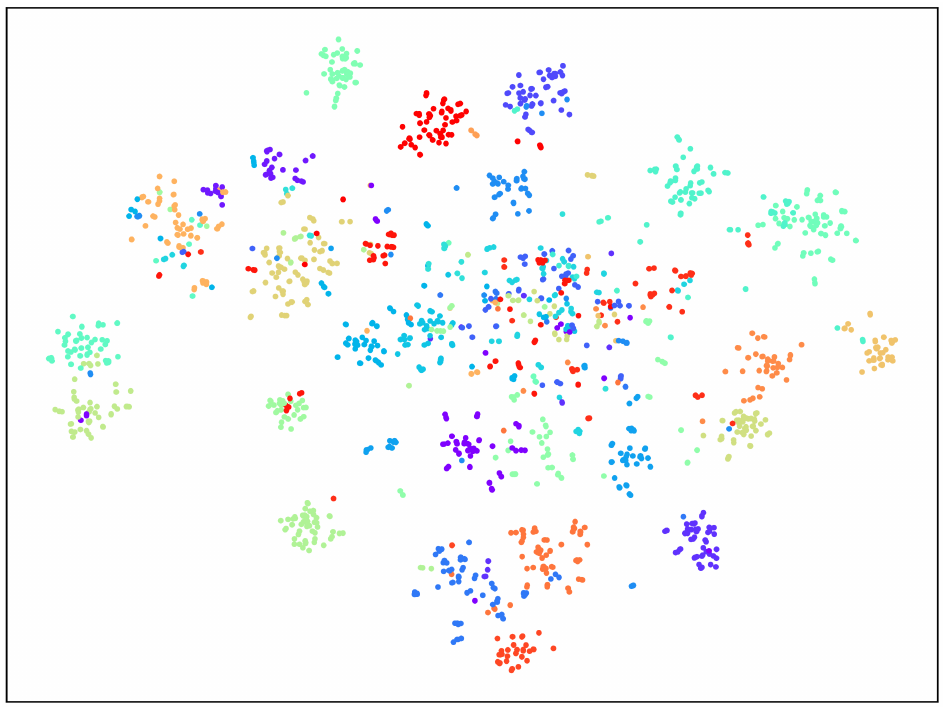}
  \includegraphics[width=0.23\textwidth]{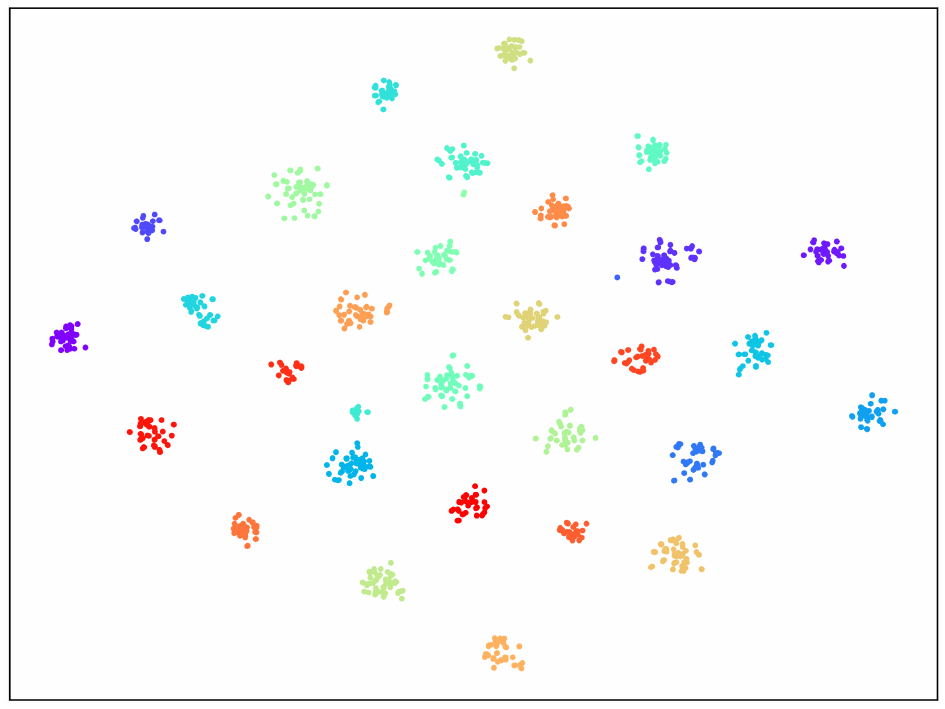}
  \\
  \makebox[0.23\textwidth]{\small (a)} 
  \makebox[0.23\textwidth]{\small (b)} 
  \\
  \caption{The visualization of high-dimensional cross-modal features. In (a) and  (b), dots with different colors represent semantic features from different scene categories.}
  \label{fig:t-sne}
\end{figure}
The t-SNE algorithm \citep{van2008visualizing} is employed to visualize high-dimensional cross-modal features. 
Figure \ref{fig:t-sne}(a) illustrates that the boundaries between different clusters are somewhat ambiguous in the baseline method CLIP-lora. As shown in Fig. \ref{fig:t-sne}(b), the proposed CMER method demonstrates an enhanced ability to differentiate semantic features across various scenes.

\subsubsection{Bad case studies}
\label{subsec: bad_case}
\begin{figure*}[!t]
    \centering
    \includegraphics [width=0.9\textwidth]{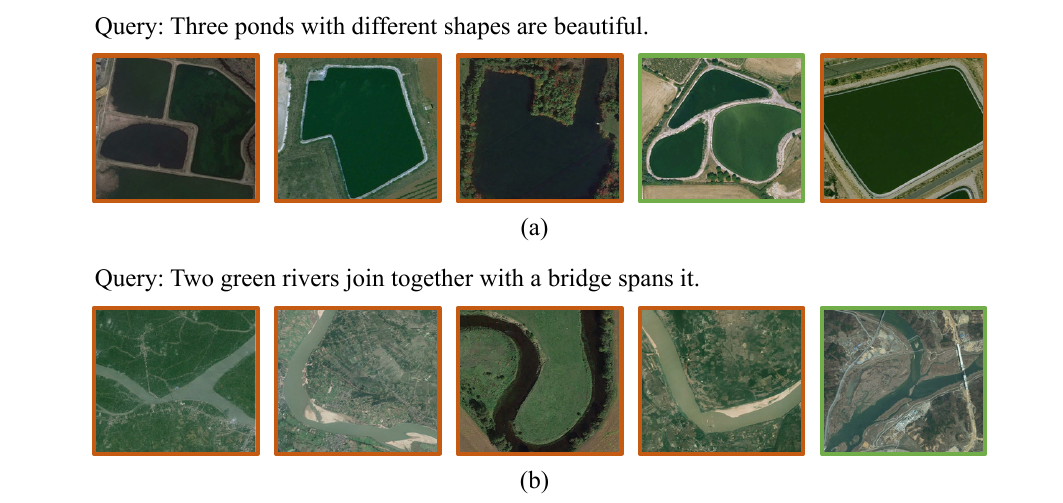}
    \caption{Failed text-image retrieval cases due to insufficient CMER capacity. The retrieved images are organized from left to right according to their similarity to the query text. The top 5 retrieved images are shown, with red boxes indicating negative matches and green boxes indicating positive matches. 
    }
    \label{fig:t2i_a}
\end{figure*}
\begin{figure*}[!t]
    \centering
    \includegraphics [width=0.92\textwidth]{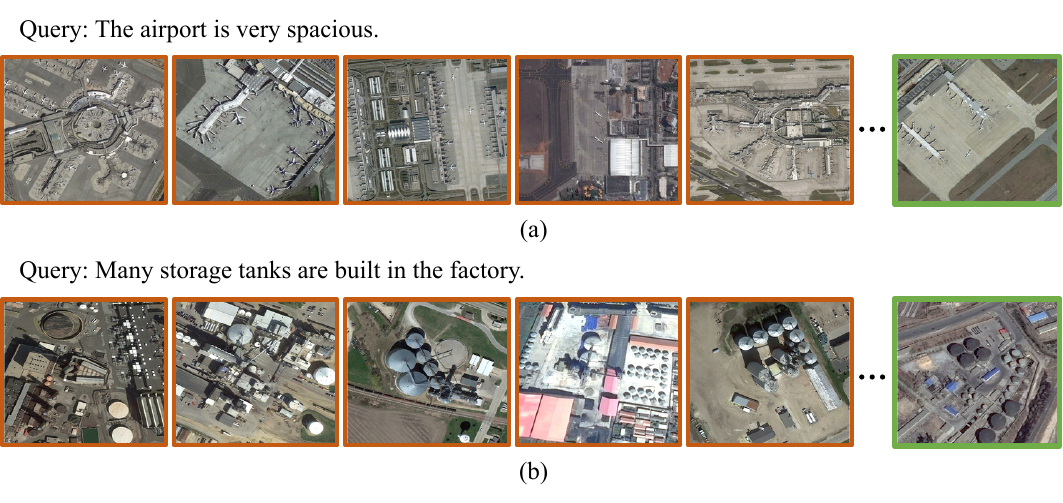}
    \caption{Failed text-image retrieval cases due to the semantic ambiguity in the RSICD. 
    }
    \label{fig:t2i_b}
\end{figure*}

To investigate the limitations of the CMER method, we specifically identify some typical failed retrieval results on the RSICD. We suggest that the performance of our method is mainly limited by modeling capacity and data quality.

As shown in Fig. \ref{fig:t2i_a}, inaccurate quantity information extraction is a noticeable weakness of CMER. When the query text explicitly states that there are three ponds. 
Based solely on the quantity information of ponds, the correct match should ideally rank first or second, rather than fourth. Figure \ref{fig:t2i_a} (b) is similar to Figure \ref{fig:t2i_a} (a), where the image at rank 1 does indeed have two rivers, but there are no bridges. 
We believe that subsequent improvements can be achieved by boosting the object-counting ability for the proposed CMER.

Dataset quality issues such as semantic concept ambiguity and high query text similarity are common challenges for RSTIR  \citep{zhang2023hypersphere}, \citep{yuan2021exploring}. 
As shown in Fig. \ref{fig:t2i_b}, we present examples where ground truth images are not among the top 5 retrieval results. According to the query text in Fig. \ref{fig:t2i_b}(a), almost all retrieved results can be described as ``spacious airports'', but the annotated ground truth does not rank in the top 5. In Fig. \ref{fig:t2i_b}(b), the first, second, and fourth RS images contain storage tanks and factory elements, satisfying the ``build-in'' relationship.  
Therefore, efforts to further improve the quality and quantity of the dataset are worthwhile, including more fine-grained, rich content description texts.
\section{Conclusion}
\label{sec:conclusion}
In this work, we propose the novel CMER for RSTIR to enhance the resource efficiency of transfer learning. The Focus-Adapter adopts the side branch structure. Its focus layer can reduce the background interference when extracting features of small targets. Moreover, the scene label prompt augmentation and the negative sample recycling strategy enhance the efficacy of data and improve the generalization performance of retrieval. Our CMER has a superior overall performance under equivalent resource constraints, which has been validated by sufficient experiments on various classical datasets. 

In our future work, we would like to boost the object counting ability of our framework, addressing the issue of mismatched quantity information. We also hope to improve the quality and quantity of the dataset, achieving fine-grained alignment of significant instances with rich content description texts. 



\bibliographystyle{cas-model2-names}

\bibliography{citation}



\end{document}